\documentclass[10pt,twocolumn,letterpaper]{article}

\usepackage[pagenumbers]{iccv} 

\usepackage{times}
\usepackage{epsfig}
\usepackage{amsmath}
\usepackage{amssymb}
\usepackage{graphicx}
\usepackage{adjustbox}
\usepackage{booktabs}
\usepackage{array}
\usepackage{subcaption}

\newcommand{\arch}[0]{\(^R\)FLAV}

\newcolumntype{?}{!{\vrule width 1.2pt}}
\newcommand{\customrule}{\specialrule{1pt}{2pt}{2pt}}

\definecolor{iccvblue}{rgb}{0.21,0.49,0.74}
\usepackage[pagebackref,breaklinks,colorlinks,allcolors=iccvblue]{hyperref}

\def\paperID{5194} 
\def\confName{ICCV}
\def\confYear{2025}

\title{ \arch: \underline{R}olling \underline{Fl}ow matching for infinite \underline{A}udio \underline{V}ideo generation}

\author{Alex Ergasti$^1$ \qquad Giuseppe Tarollo$^1$ \qquad Filippo Botti$^1$ \qquad Tomaso Fontanini$^1$ \\ Claudio Ferrari$^{1,2}$ \qquad Massimo Bertozzi$^1$ \qquad Andrea Prati$^1$ \\
$^1$ University of Parma, Department of Engineering and Architecture. Parma, Italy\\
$^2$ Univerisity of Siena, Department of Information engineering and mathematics. Siena, Italy \\
{\tt\small \{alex.ergasti, giuseppegabriele.tarollo, filippo botti, tomaso.fontanini\}@unipr.it}\\\tt\small{claudio.ferrari@unisi.it, \{massimo.bertozzi, andrea.prati\}@unipr.it} 
}

\begin{document}
\maketitle
\begin{abstract}
Joint audio-video (AV) generation is still a significant challenge in generative AI, primarily due to three critical requirements: quality of the generated samples, seamless multimodal synchronization and temporal coherence, with audio tracks that match the visual data and vice versa, and limitless video duration. In this paper, we present \arch{}, a novel transformer-based architecture that addresses all the key challenges of AV generation. We explore three distinct cross modality interaction modules, with our lightweight temporal fusion module emerging as the most effective and computationally efficient approach for aligning audio and visual modalities. Our experimental results demonstrate that \arch{} outperforms existing state-of-the-art models in multimodal AV generation tasks. Our code and checkpoints are available at \url{https://github.com/ErgastiAlex/R-FLAV}.
\end{abstract}    
\section{Introduction}
Despite extraordinary progress achieved in the field of generative AI, the development of effective joint audio-video (AV) generation approaches still represents a significant challenge. In recent years, many works have been proposed, which yet deal with just a single modality at a time, such as, for example, video \cite{weng2024art, weissenborn2019scaling, ruhe2024rolling}, or audio generation \cite{evans2024stable, liu2023audioldm, xue2024auffusion, hai2024ezaudio, fei2024fluxplaysmusic}. In addition, multimodal approaches that can shift from one modality to another have been developed as well~\cite{weng2024art, lee2023aadiff, xue2024auffusion, liu2023audioldm}, yet the task of simultaneous AV generation remains quite unexplored due to the significant challenges it encompasses. Other than the clear technical difficulties, we identified three major critical requirements that an effective AV generation system should comply with: \textit{(i)} audio and video quality, \textit{(ii)} seamless multimodal synchronization and temporal coherence, \textit{(iii)} infinite video length. In other words, the duration of the generated video should not be constrained by a fixed length, while ensuring good quality without producing degenerate results and audio tracks that consistently match the dynamics of the visual data. Although there have been recent proposals in this field, they all struggle to meet all the above features.

Among the few works that addressed this task, MM-Diffusion (MMD)~\cite{ruan2023mm} pioneered a novel fusion mechanism that allowed joint AV generation. However, such a solution is restricted in terms of video duration, which is fixed in length during the diffusion process.  
Following the introduction of transformers in diffusion model architectures \cite{peebles2023scalable}, Kim \etal \cite{kim2024a} proposed a novel method to generate longer videos. In order to do so, an auto-regressive approach is needed, where the generation of a new sequence of frames is conditioned by the last generated frames of the previous sequence. Even though this approach enables the generation of videos with a variable number of frames, in the long run the quality may deteriorate because of the error accumulation that affects auto-regressive kind of strategies. 

A solution to this problem has been proposed by Ruhe \etal~\cite{ruhe2024rolling} by introducing Rolling Diffusion, a novel technique for training diffusion models for video generation that uses a sliding window denoising process. It is based on the idea that, in a sequence, the prediction of frames that are distant in the future is more uncertain with respect to temporally-closer ones. Thus, the amount of noise of the diffusion process for each frame should be proportional to the temporal distance. This approach resulted preferable to the auto-regressive one for improving the quality and consistency of longer sequences, although it does not account for the generation of synchronized audio tracks.  

In light of the above, in this paper we propose a novel architecture for joint AV generation, called \arch, specifically designed to address the key challenges discussed so far. In general, our method demonstrates better performance in comparison to previous approaches in terms of generation quality and AV consistency.   
Above all though, our model takes a significant step forward by enabling \textbf{infinite AV generation}. Our solution allows for the generation of paired AV sequences \textbf{without any restriction on their duration}, while maintaining high multimodal synchronization and quality.
To achieve these results, we carefully designed several components in our architecture. In particular, we revisit the way audio and video are encoded so as to adapt the multimodal stream to be processed via rolling diffusion mechanism and we propose a new lightweight cross-modality interaction module. Ultimately, our contributions are as follows:
\begin{itemize}
    \item A rolling flow matching model supporting infinite AV generation with sustained visual and acoustic consistency.
    \item An exploration of different alternatives for cross-modality interaction, resulting in a lightweight AV fusion module which does not require attention mechanism.
\end{itemize}

\noindent Our code and models will be \textbf{fully open source}.
\section{Related Work}
\paragraph{Multimodal Generation.}
Multimodal models have the goal of learning a more general and comprehensive representation of multiple data modalities, such as video, sound, and text. In particular, for these tasks several key challenges have to be faced, such as temporal coherence for time-dependent data and modality alignment, but also more practical ones like computational performance. 
Typically, previous studies focused on single-modality generation \cite{gong2021ast, ruhe2024rolling, weissenborn2019scaling} or text-to-video, text-to-audio, and other conditional generation \cite{evans2024stable, ge2022long, ghosal2023text, liu2024autoregressive, weng2024art, xue2024auffusion}, which are common applications for generative models. 
%
%
Among these, the most successful ones are based on diffusion models, but despite their versatility, the number of applications in multimodal tasks remains limited and lacks variety, due to the intrinsic complexity of such problems.

Nevertheless, an application that has gained popularity is joint audio-video generation, which recently saw a surge of different approaches based on diffusion models \cite{ruan2023mm, wang2024av, xing2024seeing, yang2023cmmd, tang2023any, kim2024a}. Among these, MM-Diffusion (MMD) \cite{ruan2023mm} proposed a U-Net based architecture, featuring two separate branches for audio and video, merged by a random-shift attention block, which allowed sound and visual data to influence and synchronize with each other. Upon MMD, AV-DiT \cite{wang2024av} exploited a shared DiT-XL/2 backbone \cite{peebles2023scalable}, pre-trained on ImageNet, by introducing trainable adapter layers for audio and video, demonstrating better alignment and generation of modality.
Finally, Kim \etal \cite{kim2024a} introduced a novel parametrization of the diffusion timestep for the forward process, applying a different diffusion timestep across each modality and temporal dimension. This approach, combined with their model architecture, is capable of surpassing previous work. However, the output produced is limited to 2.125 seconds, and an autoregressive mechanism is still required to produce video longer than 2.125 seconds, with the risk of error accumulation and performance degradation.

Our method improves upon such models by exploiting a rolling approach to generate high-quality, coherent, long-duration videos with aligned audio. Moreover, we propose a novel strategy to fuse and align audio and video during training through a lightweight temporal cross-modality interaction block. This design also allows our model to perform video-to-audio and audio-to-video tasks, along with joint audio-video generation. Finally, since our model does not employ any video or audio encoder, input and output lengths are not constrained to be multiples of the encoder length, allowing training with an arbitrary number of frames and the generation of videos with unconstrained duration.

\paragraph{Diffusion Models.}
Diffusion models have achieved great success in various tasks due to their versatility and scalability \cite{blattmann2023stable, kong2020diffwave, liu2023audioldm, ruan2023mm, wang2024av} and have presented numerous backbones for different applications \cite{rombach2022high, peebles2023scalable, bao2023all}. A drawback of diffusion models though is that to generate new samples, the traditional diffusion process involves a complex backward path that requires multiple integration steps. To address this, a method called Rectified Flow, or Flow Matching, was proposed \cite{liu2022flow, lipman2023flow} to learn a straight path between data and noise by linearly interpolating them, demonstrating improved performance with appropriate weight adjustments compared to standard diffusion~\cite{esser2024scaling}. Furthermore, Ruhe \etal \cite{ruhe2024rolling} proposed a sliding window approach for video generation called Rolling Diffusion. In this method, model inputs consist of frames with progressively increasing diffusion steps, where frames present higher noise levels depending on their position in time, with the last frame consisting of full noise. During each diffusion step, frames are incrementally denoised until the first frame in the window is fully denoised, at which point it is removed from the window and replaced with a new noise frame at the end. This approach effectively addresses the challenge of maintaining spatial and temporal consistency in video generation.

In our work, we adapt this method to audio-video generation by pairing it with a transformer architecture as backbone of our model, following \cite{bao2023all, esser2024scaling, peebles2023scalable}, and using flow matching as model framework.

\section{Methodology}
Our proposed architecture, called \arch, is a rolling rectified-flow model designed for AV generation. More in detail, we modified the rolling diffusion methodology \cite{ruhe2024rolling} to enable training using rectified flow matching \cite{lipman2023flow, liu2022flow}. 

\subsection{Architecture}

\begin{figure}[h!]
    \centering
    \includegraphics[width=\linewidth]{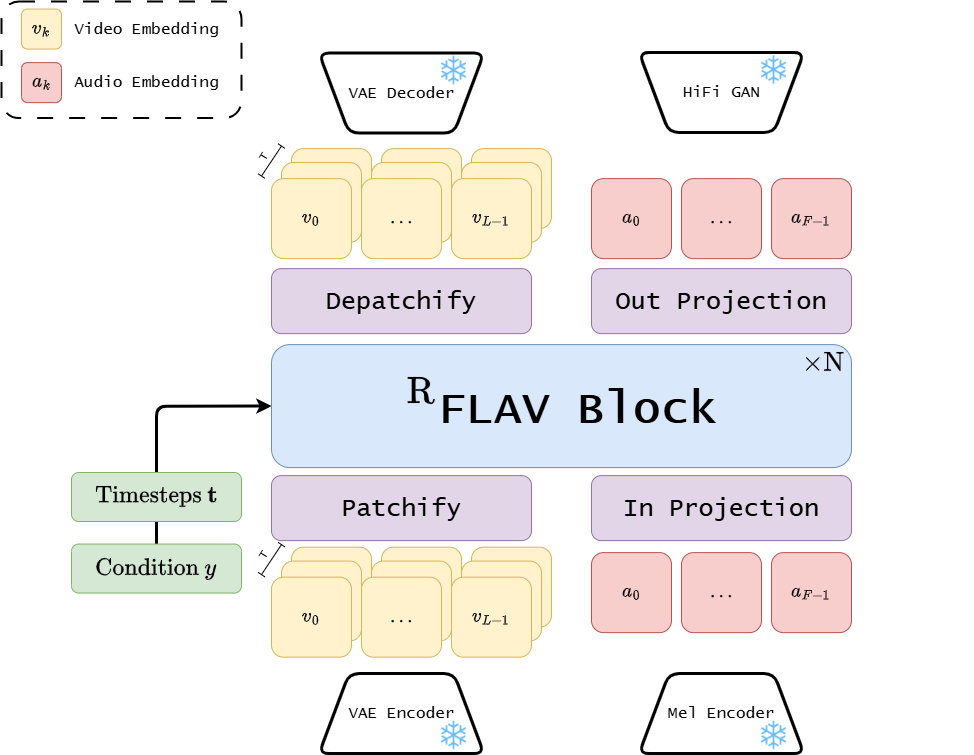}
    \caption{An overview of our \arch{} model architecture.} 
    \label{fig:architecture}
\end{figure}

Our architecture (see Fig.~\ref{fig:architecture}), called \arch, is a transformer-based model designed to allow AV generation of any length. This is possible since our model does not depend on any audio or video encoder, which typically impose generated videos to be multiples of their output sizes. 

The model processes both video and audio in two separate branches. Fusion of different distributed modalities is avoided at early stages; they are instead combined later in the architecture. This allows for an early \textit{intra-modality} interaction, performed with self-attentions, before the inter-modality interaction. 

\begin{figure}[h!]
    \centering
    \includegraphics[width=0.9\linewidth]{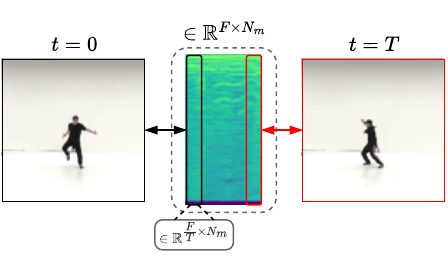}
    \caption{Temporal alignment between video frames and mel spectrogram segments. Each video frame corresponds to a fixed-size section (F/T) of the mel spectrogram, allowing for a 1:1 mapping.}
    \label{fig:mel-to-frame}
\end{figure}

\paragraph{Video and Audio encoding.}
To generate videos of any length, we designed our system to synthesize samples frame-by-frame. Our approach uses an image encoder, the same employed in single image generation \cite{rombach2022high}, instead of a video encoder to reduce the spatial dimension of each frame of the video, avoiding any temporal compression. The output of the encoder is a video $\mathbf{v}\in\mathbb{R}^{T\times c\times h\times w}$, where $T$ is the temporal dimension, $c$ represents the number of encoder channels and $h=H/f$, $w=W/f$, with $H$ and $W$ being the original size of the video and $f$ being the downsampling factor. The audio $\mathbf{a}\in\mathbb{R}^{F\times N_m}$ is instead represented by its mel spectrogram obtained from the raw waveform, where $F$ is the number of time frames and $N_m$ the number of mel bins. 
Since each modality is encoded without performing any temporal compression (\textit{i.e.}, without any video or audio encoder), a one-to-one reference between audio and video frames can be established. More in detail, each video frame can be associated with the corresponding part of the mel spectrogram, having size $F/T$, for a better interaction between the two modalities. This is shown in Fig.~\ref{fig:mel-to-frame}. 

\begin{figure*}[t!]
    \centering
    \begin{subfigure}{0.33\textwidth}
        \includegraphics[width=\textwidth]{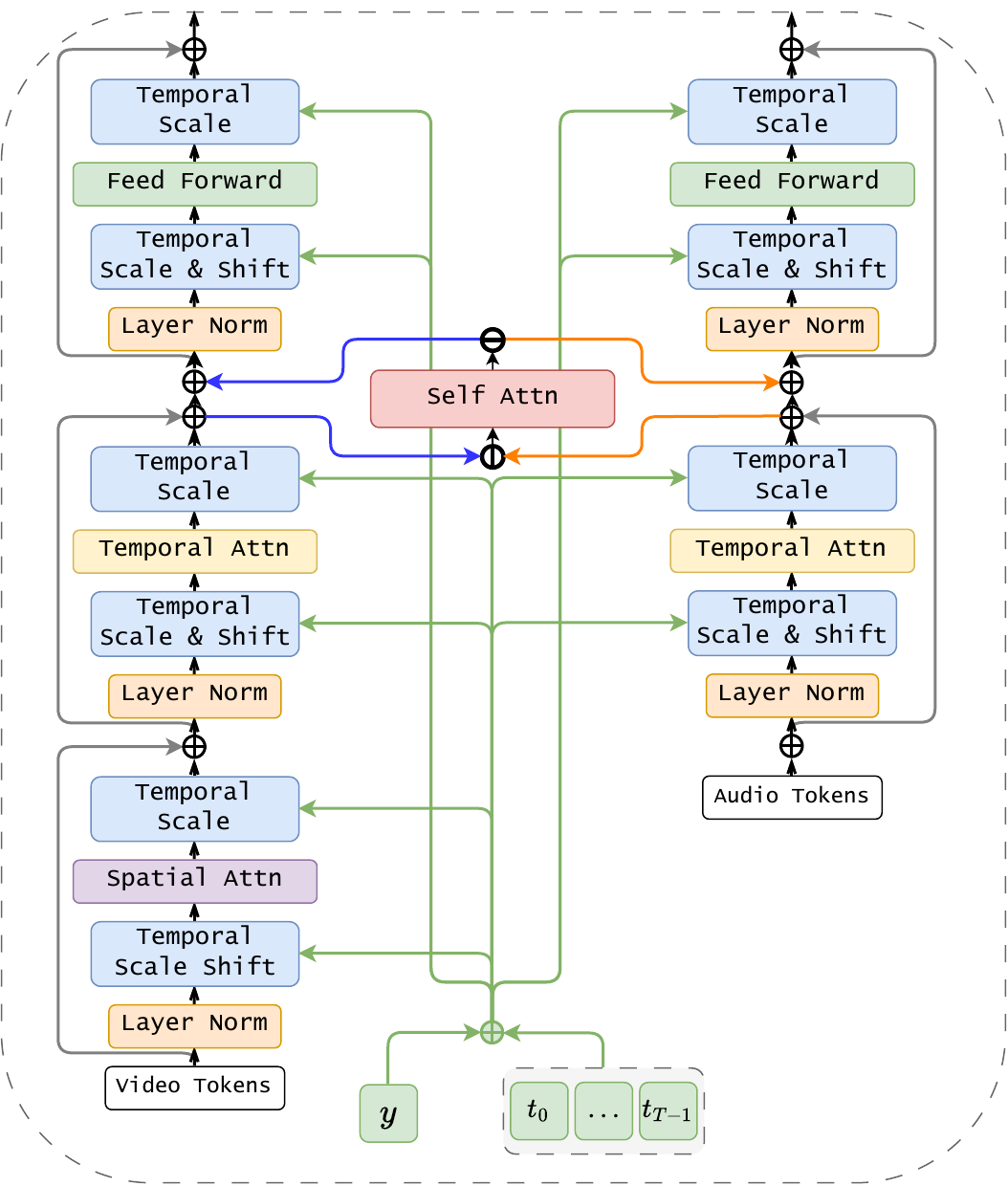}
        \caption{}
        \label{fig:block3}
    \end{subfigure}
    \begin{subfigure}{0.33\textwidth}
        \includegraphics[width=\textwidth]{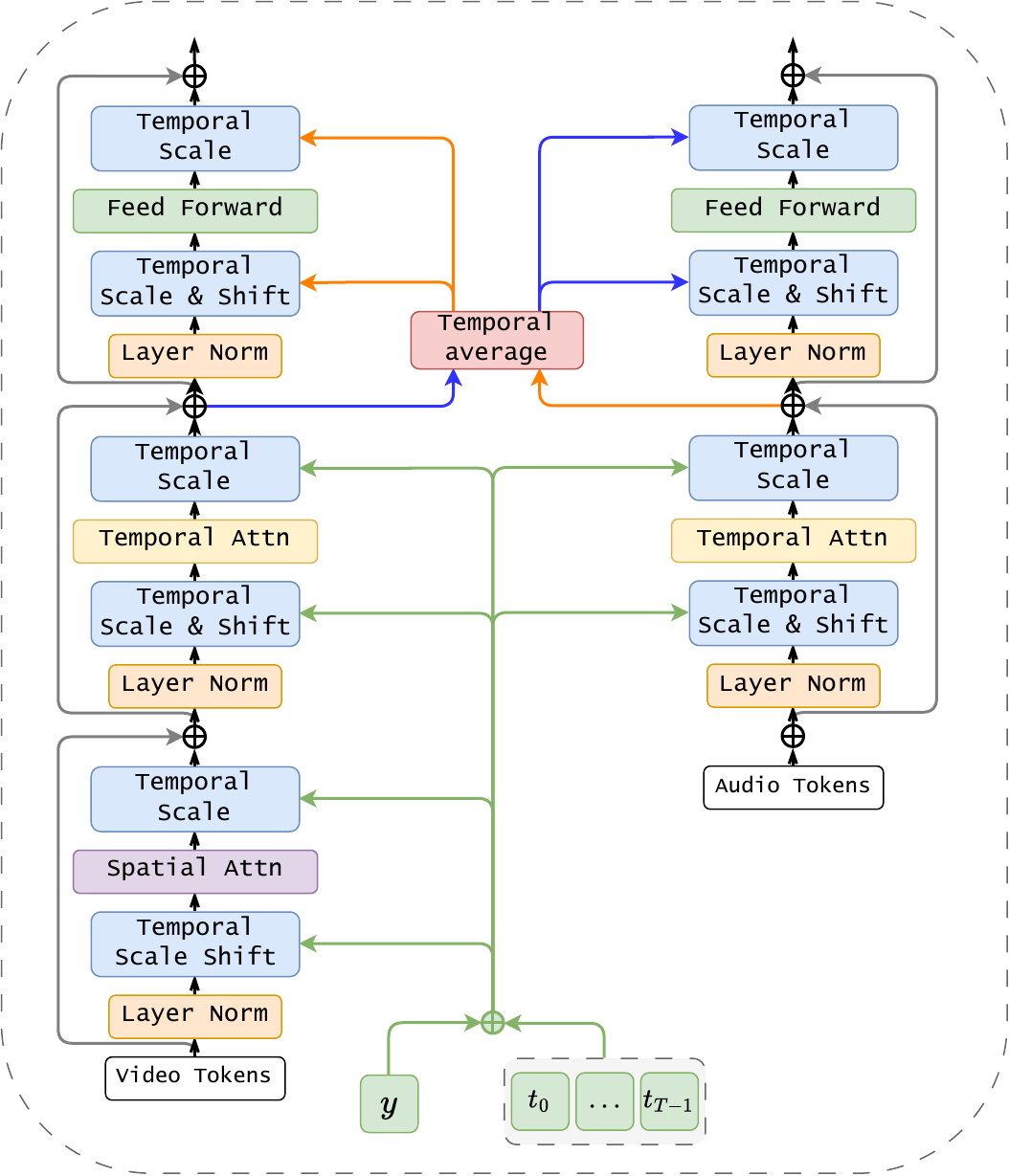}
        \caption{}
        \label{fig:block1}
    \end{subfigure}
    \hfill
    \begin{subfigure}{0.33\textwidth}
        \includegraphics[width=\textwidth]{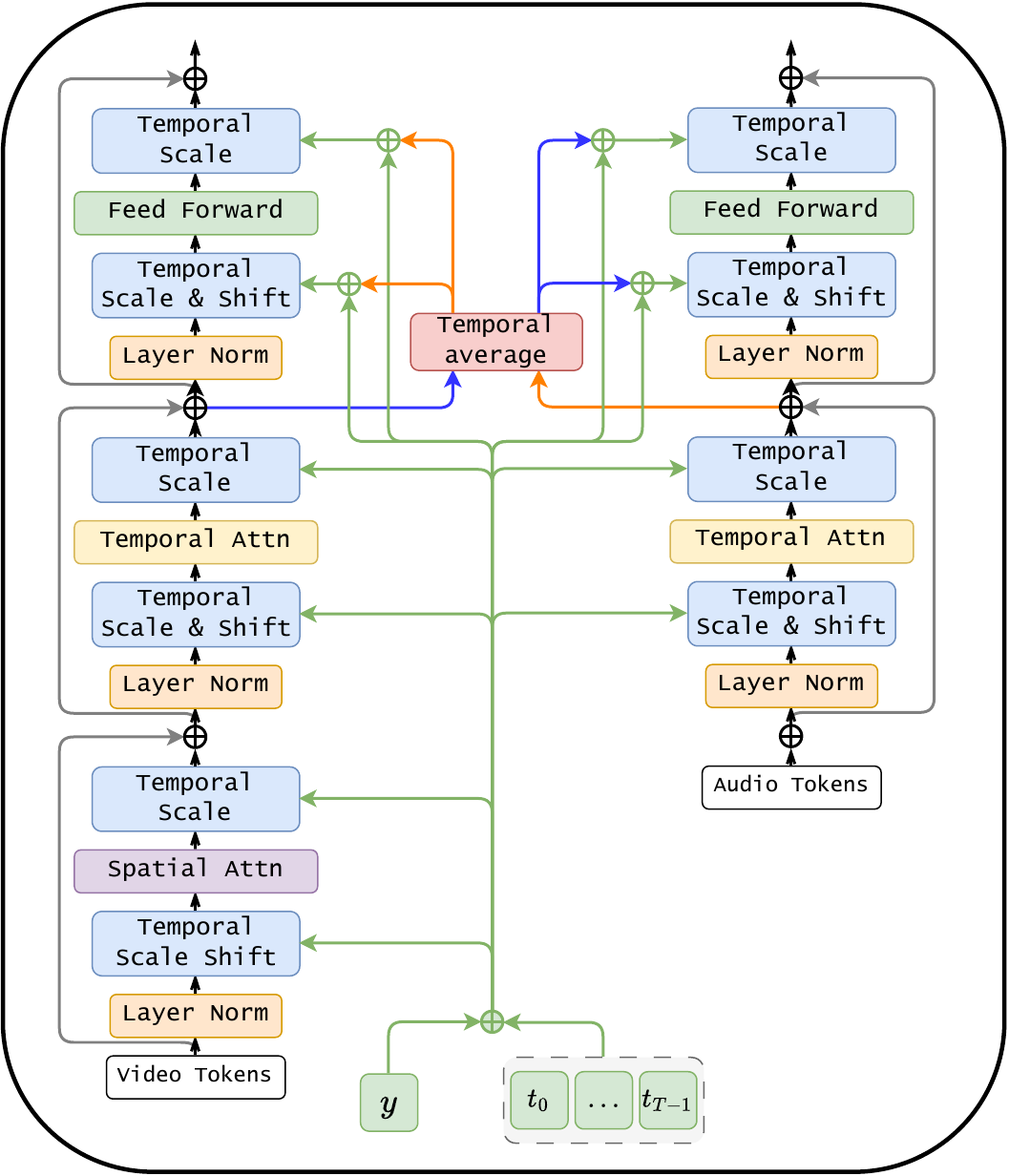}
        \caption{}
        \label{fig:block2}
    \end{subfigure}
    \hfill

    \caption{a) Cross-modal interaction via self-attention, where  \rotatebox[origin=c]{90}{$\ominus$} and $\ominus$ mean concatenation and split. b) Lightweight cross-modality interaction mechanism with temporal average modulation. c) Our final proposed \arch{} block, an enhanced lightweight mechanism incorporating timestep embedding $t$ and optional class conditioning embedding $c$.}
    \label{fig:blocks}
\end{figure*}

The processing occurs in parallel for both modalities. The video passes through a patchify layer, producing a feature embedding tensor $\mathbf{v}\in\mathbb{R}^{T\times L\times D}$, where $L$ is the number of spatial patches per frame (computed as $L=\frac{h\times w}{p^2}$, with frame dimensions $h,w$ and patch size $p$) and $D$ is the hidden size. Simultaneously, the audio undergoes a linear projection to obtain $\mathbf{a}\in\mathbb{R}^{F\times D}$, where $F$ is the number of audio segments and $D$ matches the hidden dimension of the video, resulting in $T\times L$ video patches and $F$ audio patches. 

Thanks to its design, the proposed architecture enables the generation of videos of any length, whereas with the usage of a video (or audio) encoder, the length must be a multiple of the encoder's output size.

\paragraph{\arch{} block.}

The main block of our architecture is divided into two parallel branches, one for the video and one for the audio. The video branch sequentially applies spatial and temporal attention. The audio branch applies only temporal attention since the audio has no spatial structure.

Both branches incorporate time-dependent feature modulation before and after each attention layer. This is performed with a custom version of DiT adaptive layer normalization (AdaLN) \cite{peebles2023scalable}, where each video frame and the corresponding part of the audio are modulated by a different timestep embedding $t$ (see Sec.~\ref{sec:rf_matching}) and, optionally, by a timestep-invariant class conditioning embedding $y$. To perform cross-modality fusion we propose and explore 3 different alternative blocks structure, as shown in Fig.~\ref{fig:blocks}.

The first block (Fig.~\ref{fig:block3}) implements cross-modal interaction via self-attention. Audio embeddings are reshaped to $\mathbb{R}^{T\times F/T\times D}$ and concatenated with video embeddings, obtaining $\mathbf{av}\in\mathbb{R}^{T\cdot(L+F/T)\times D}$.
The feature map is then processed by causally masked self-attentions. Finally, the result is split again into the corresponding video and audio embeddings and summed to the respective branch.
Although this approach offers sophisticated cross-modal interaction, it requires much higher computational and memory costs w.r.t. the following blocks.

The second block (Fig.~\ref{fig:block1}) employs a lightweight modulation mechanism that starts by computing temporal averages. More in detail, the video features $\mathbf{v}\in\mathbb{R}^{T\times L\times D}$ are reduced to $\overline{\mathbf{v}} \in \mathbb{R}^{T\times 1 \times D}$, while audio features $\mathbf{a}\in\mathbb{R}^{F\times N_m}$ are first reshaped to $\mathbb{R}^{T\times \frac{F}{T}\times D}$ and then reduced to $\overline{\mathbf{a}} \in \mathbb{R}^{T \times 1 \times D}$. Next, $\overline{\mathbf{v}}, \overline{\mathbf{a}}$ are used for cross-modal fusion. Before and after the feedforward layer in each branch, we apply the modulation operations used for the timesteps, but parameterized by the averaged features of the other branch.

The final block (Fig.~\ref{fig:block2}) is built upon the second block. It sums the timestep embedding $t$ and the (optional) class embedding $y$ to the temporal average embedding. This enhances the propagation of timestep and class information.

Furthermore, it is worth noting that in all the proposed blocks, cross-modality interaction is strategically placed after attention layers and before feed forward layers, allowing each modality branch to independently process information through its respective attention mechanisms before influencing the other modality branch.

\subsection{Flow Matching}
Flow matching defines the forward path as a linear trajectory between a sample $x$ drawn from the data distribution and noise $\epsilon$ obtained from a Gaussian distribution $\mathcal{N}(0,1)$, where the trajectory depends on a time parameter, represented by $t$, with $t$ ranging from 0 to 1.
\begin{equation}
    x_t = tx+(1-t)\epsilon
    \label{eq:trajectory}
\end{equation}
The model is then trained to predict the velocity vector $\phi$, which drives a sample from noise to the data distribution. The velocity vector $\phi$ is defined as the derivative of the trajectory path with respect to $t$: \begin{equation}
    \phi=\frac{dx_t}{dt}=(x-\epsilon)
\end{equation}
Hence, the model predicts the velocity vector $\hat{\phi}(x_t,t)$ taking as input both $x_t$ and the timestep $t$. The final loss is:
\begin{equation}
\adjustbox{width=0.90\linewidth}{
    $
    \mathcal{L}_\mathrm{FM}(x)=\mathbb{E}_{t\sim\mathcal{U}(0,1), \epsilon\sim\mathcal{N}(0,1)}\lambda_t||(x-\epsilon)-\hat{\phi}(x_t,t)||^2
    $   
}
\label{eq:loss_fm}
\end{equation}
The loss is scaled by a weight factor $\lambda_t$. In \cite{esser2024scaling}, the authors show that the best factor is obtained with $\lambda_t=\text{logit-normal}(t;0,1)$.

\subsection{Rolling Flow Matching}\label{sec:rf_matching}

\begin{figure*}
    \centering
    \begin{subfigure} {0.595\textwidth}
        \includegraphics[width=\textwidth]{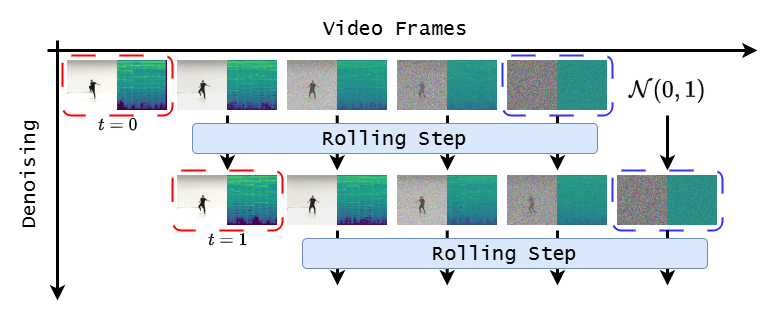}
        \caption{}
        \label{fig:rolling}
    \end{subfigure}
    \hfill
    \begin{subfigure}{0.40\textwidth}
        \includegraphics[width=\textwidth]{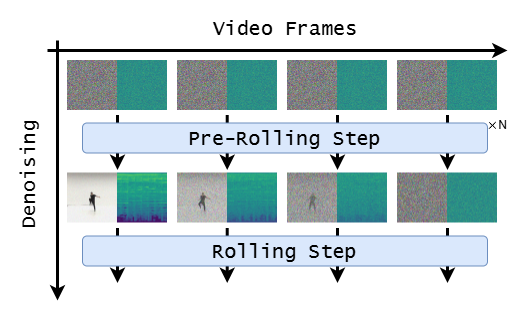}
        \caption{}
        \label{fig:pre-rolling}
    \end{subfigure}
    \caption{a) Rolling phase: at each step, a new clean frame is produced (highlighted in red) and subsequently removed from the window. Then, a new noisy frame, (highlighted in blue), is appended to the end of the window. b) Pre-rolling phase: the frames are gradually denoised starting from a full noise configuration. The pre-rolling phase goes on for $N$ steps, until the window is ready for the rolling phase.}
\end{figure*}

Rolling Diffusion Model, introduced by Ruhe \etal \cite{ruhe2024rolling}, serves as a foundational framework for generating infinite video sequences without necessitating an auto-regressive approach. More in detail, it employs a sliding window denoising technique, in which each frame corresponds to a distinct, but progressively higher, denoising timestep. Once the initial frame is fully denoised, the entire window shifts to allow a new noisy frame at the end of the window. In our architecture, we propose a novel rolling model which simultaneously generates both video and audio, without any pre-fixed length and with substained quality.


Given $\mathbf{v}\in\mathbb{R}^{T\times c\times h\times w}$ and $\mathbf{a}\in\mathbb{R}^{T \times F/T \times N_m}$, we define the forward process to obtain the noisy video $\mathbf{\hat{v}}$ and the noisy audio embeddings $\mathbf{\hat{a}}$ as:
\begin{equation}
\hat{v}_k = t_kv_k + (1-t_k)\epsilon_{v,k}, \quad \epsilon_{v,k}\sim\mathcal{N}(0,1)
\end{equation}
\begin{equation}
\hat{a}_k = t_ka_{k} + (1-t_k)\epsilon_{a,k}, \quad \epsilon_{a,k}\sim\mathcal{N}(0,1)
\end{equation}
where $t_k$ is the timestep adjusted for each frame $k\in[0,T-1]$ and its corresponding $F/T$ mel spectrogram vector. 

Following \cite{ruhe2024rolling}, we introduce two distinct temporal scheduling mechanisms for $t_k$ depending on a variable $w\sim\mathcal{U}(0,1)$. The rolling phase scheduler, $t_k^r(w)$, is designed for continuous generation in which the model produces content sequentially, while the pre-rolling phase scheduler, $t_k^p(w)$, is specifically designed to handle the initial state where both modalities start as pure noise. 

During the rolling phase (see Fig. \ref{fig:rolling}) the timestep for each window frame $k$ is computed as:
\begin{equation}
    t_k^r(w)=1-\frac{k+w}{T}
\end{equation}
\noindent with each $t^r_k(w)\in[0,1]$ having a distance between each consecutive timesteps of $1/T$ and $t^r_k(w)<t^r_{k+1}(w)$.

However, using only the rolling phase timesteps formulation would result in $t^r_k(w)$ never being 0 for all the frames in the sequence, meaning that it would not be possible to sample new AV pairs from pure noise. Thus, a pre-rolling phase (see Fig.~\ref{fig:pre-rolling}) is added during training, defining an alternative timestep formulation to allow the model to handle fully noisy inputs. The computation becomes:
\begin{equation}
    t_k^p(w)=\mathrm{clamp}\left(1-\left(\frac{k}{T}+w\right);0,1\right)
\end{equation}
This ensures that $t^p_k(w)\in[0,1]$, but with $t_k^p(w)\le t^p_{k+1}(w)$. 
Specifically, $w=1$ represents the initial state when all frames are pure noise. Then, moving towards $w=0$, the noise is gradually removed from each frame in the sequence depending on its position in time. Finally, when $w=0$, the first frame will be noise-free and the last one will be pure noise and the rolling phase can begin. The final timestep vector $\mathbf{t}(w)=[t_0(w),\dots,t_{T-1}(w)]$ becomes:
\begin{equation}
\mathbf{t}(w) =
\begin{cases}
t^p(w), & \text{with probability } \theta \\
t^r(w), & \text{with probability } 1 - \theta
\end{cases}
\end{equation}

\noindent where $\theta \in [0,1]$ is an hyperparameter controlling the balance between pre-rolling and rolling phase during training.

The final model loss is then formulated from Eq.~\ref{eq:loss_fm} as:
\begin{equation}
\adjustbox{width=0.85\linewidth}{
$
    \mathcal{L}(\mathbf{v},\mathbf{a})=\mathbb{E}_{\substack{w\sim\mathcal{U}(0,1)\\ \mathbf{\epsilon}_v\sim\mathcal{N}(0,1)\\\mathbf{\epsilon}_a\sim\mathcal{N}(0,1)}}
    \left[\mathcal{L}_\mathbf{v}(\mathbf{v},\mathbf{t}(w),\mathbf{\epsilon}_v)+ \mathcal{L}_\mathbf{a}(\mathbf{a},\mathbf{t}(w),\mathbf{\epsilon}_a)\right]
$
}
\end{equation}
where $\mathcal{L}_\mathbf{v}$ and $\mathcal{L}_\mathbf{a}$ are defined as:
\begin{equation}
\adjustbox{width=0.85\linewidth}{
$
\begin{aligned}
    \mathcal{L}_\mathbf{v}(\mathbf{v},\mathbf{t}(w), \mathbf{\epsilon}_v)=\left[
    \frac{1}{T} \sum_{k=0}^{T-1} \lambda_{t_k} \left( ||(v_k-\epsilon_{v,k}) - \hat{\phi}(\hat{v}_k, t_k(w))||^2
    \right) \right] \\
    \mathcal{L}_\mathbf{a}(\mathbf{a},\mathbf{t}(w), \mathbf{\epsilon}_a)= \left[
    \frac{1}{T} \sum_{k=0}^{T-1} \lambda_{t_k} \left(||
    (a_k-\epsilon_{a,k}) - \hat{\phi}(\hat{a}_k, t_k(w))||^2 
    \right) \right]
\end{aligned}
$
}
\end{equation}
with $\lambda_{t_k} = \text{logit-normal}(t_k; 0, 1)$.


\section{Experiments}

\subsection{Model details}

Our model is composed of 12 \arch{} blocks and a window size of 10 frames. To encode each video frame we employ the stable diffusion encoder \cite{rombach2022high}, which has a downsize factor $f=8$ and with $c=4$. For audio, we use the same vocoder as \cite{xue2024auffusion}, with $N_m=256$. $\theta$ is fixed to 0.2.

\subsection{Datasets}
We follow previous work \cite{ruan2023mm} to evaluate our model. Specifically, we train our model on two datasets, Landscape~\cite{ruan2023mm} and AIST++~\cite{li2021ai}. Landscape dataset contains 9 different settings (\textit{i.e.}, explosion, fire crackling, raining, splashing water, squishing water, thunder, underwater bubbling, underwater burbling, and wind noise) providing 2.7 hours of high-quality audio-video pairs. AIST++~\cite{li2021ai} is a subset of AIST \cite{tsuchida2019aist} that provides 5.2 hours of video featuring paired audio and dancer movements. 

\subsection{Evaluation metrics} \label{subsec:metrics}
We evaluated ours and other state-of-the-art models with 2048 generated samples at $64\times 64$ resolutions and 16 frames per video sample. Following previous work \cite{ruan2023mm} we used Fréchet Video Distance (FVD) and Kernel Video Distance (KVD) to measure video quality, employing the I3D video classifier pre-trained on Kinetics-400~\cite{carreira2017quo}. To measure audio quality, we used Fréchet Audio Distance (FAD) calculated from a pre-trained AudioCLIP classifier \cite{guzhov2022audioclip}.

\subsection{Ablation Study}
In Table~\ref{tab:ablation-blocks}, we present a comparison of all our proposed blocks on the AIST++ dataset. All the metrics are calculated using 20 denoising steps. Indeed, the proposed block (c) outperforms blocks (a) and (b) in terms of overall performance. Specifically, compared to block (b), our architecture achieves a more effective propagation of the timestep embedding, leading to improved performance in the diffusion process. Additionally, compared to block (a), our proposed block is more efficient and faster, requiring fewer computational resources and less memory.
In addition, we also test a window size of 5 frames and 20 frames. We observe, as reported in Table~\ref{tab:ablation-ws}, that 10 frames is the optimal solution. The reason is that a smaller window (e.g., 5 frames) may not capture enough temporal context, leading to incomplete motion patterns while a larger window can introduce redundant information and increase noise. The 10-frame window provides the right balance, ensuring sufficient context while avoiding excessive complexity.


\begin{table}[h!]
    \centering
    \adjustbox{width=\linewidth}{
    \begin{tabular}{c|cc?ccc}
        \toprule
        \multicolumn{6}{c}{\textbf{AIST++}} \\
        \midrule
        Model & Time (s) & Memory Usage (GB) & FVD $\downarrow$ & KVD $\downarrow$ & FAD $\downarrow$ \\ 
        \midrule
        \textbf{\arch}-(a) & 15.47 & 7.9 & 53.38 & 7.33 & 8.70 \\ 
        \textbf{\arch}-(b) & 3.92 & 2.7 & 51.31 & 9.76 & \textbf{8.30} \\ 
        \textbf{\arch}-(c) & 3.94 & 2.7 & \textbf{50.92} & \textbf{8.73} & 8.40 \\
        \bottomrule
    \end{tabular}}
    \renewcommand{\thetable}{\arabic{table}.A}
    \caption{Comparison between inference time for a sample of 16 frames, memory usage and quantitative metrics of all the 3 proposed blocks. Time was calculated on a NVIDIA RTX 4090 gpu.}
    \label{tab:ablation-blocks}
    \adjustbox{width=\linewidth}{
    \begin{tabular}{c|c?ccc}
        \toprule
        \multicolumn{5}{c}{\textbf{AIST++}} \\
        \midrule
        Model & Window Size & FVD $\downarrow$ & KVD $\downarrow$ & FAD $\downarrow$ \\ 
        \midrule
        \textbf{\arch}-(c) & 5  & 65.42 & 10.41 & 8.46 \\
        \textbf{\arch}-(c) & 10 & \textbf{50.92} & \textbf{8.73}  & \textbf{8.40} \\
        \textbf{\arch}-(c) & 20 & 77.55 & 12.71 & 8.53 \\
        \bottomrule
    \end{tabular}}
    \addtocounter{table}{-1}
    \renewcommand{\thetable}{\arabic{table}.B}
    \caption{Comparison of \arch{}-(c) with different window size.}
    \label{tab:ablation-ws}
\end{table}

\subsection{Comparison with SOTA models}
We compare our model with the current SOTA models \cite{ruan2023mm, kim2024a, wang2024av}, performing both a quantitative and qualitative analysis. For open source models, such as \cite{ruan2023mm}, we recompute the metrics on our hardware, and, vice versa, for closed source models (\cite{kim2024a, wang2024av}), we took the metrics directly from the original papers. The reason why, when possible, we choose to calculate the metrics from the SOTA models again is that, as pointed out in \cite{parmar2022aliased}, the way images are resized and compressed can have a large impact on the common evaluation metrics of generative models. Therefore, by reproducing the same settings for all the open-source generative models we compare with, we ensure a fair comparison. Unfortunately, this cannot be done for closed-source models for obvious reasons. For the sake of a fair comparison, re-implementing the methods is not a viable solution since both \emph{(i)} fine-grained technical details are not provided in the related papers and exactly reproducing the implementation is almost impossible and \emph{(ii)} would likely lead to inconsistent results. Thus, we opted for copying the numbers.

\begin{table}[h!]
    \centering
    \adjustbox{width=\linewidth}{
\begin{tabular}{c|cc?ccc?ccc}
    \toprule
     & & & \multicolumn{3}{c?}{\textbf{AIST++}} & \multicolumn{3}{c}{\textbf{Landscape}} \\ 
     \midrule
     Model & Params & Steps & FVD $\downarrow$ & KVD $\downarrow$ & FAD $\downarrow$ &  FVD $\downarrow$ & KVD $\downarrow$ & FAD $\downarrow$ \\ \midrule
     GT & - & - &  3.67 & -0.28 & 8.25 & 7.45 & -0.15 & 9.33
 \\
     \midrule
     MMD \cite{ruan2023mm} & 426 M & 25 & 224.14 &  50.52 & 11.41 & 177.29 & 7.63 & 9.73 \\ 
     Wang \etal$^*$ \cite{wang2024av} & 931 M & 250 & 68.88 & 21.01 & 10.17 &  172.69 & 15.41 & 11.17 \\
     \customrule
     \textbf{\arch} & 421 M  & 20 & 50.92 & 8.73 & 8.40 & 86.53 & 3.36 & 10.49  \\ 
     \textbf{\arch} & 421 M & 100 & 38.93 & 6.58 & 8.35 & 85.44 & 3.73 & \textbf{9.61}  \\ 
     \textbf{\arch} & 421 M & 200 & \textbf{38.36} & \textbf{6.15} & \textbf{8.28} & \textbf{80.19} & \textbf{3.30} & 9.88  \\ 
     \bottomrule
    \end{tabular}
    }
    \caption{A comparison between our method and the current SOTA models, calculated with 2048 samples at $64\times64$ resolution. $^*$Metrics of the model are taken from the paper since \textbf{neither code nor checkpoints are available.}
    }
    \label{tab:results}
\end{table}

\paragraph{Quantitative analysis.}
The results of the SOTA models \cite{ruan2023mm, kim2024a, wang2024av} are shown in Table.~\ref{tab:results}. 
Furthermore, the comparison with  AVDiT by Kim \etal \cite{kim2024a} is presented in Table~\ref{tab:nips-comparison}. The reason for a different table is that, for AIST++ and Landscape datasets, AVDiT by Kim \etal presented results only for the tasks of Audio2Video (A2V) and Video2Audio (V2A) generation, in which original audio is used to generate a video and vice versa, and not for fully joint AV generation. Additionally, new results could not be produced since neither code nor checkpoints are currently available. Therefore, we took their FVD and KVD from A2V generation and their FAD from V2A generation and compared them with our results for these tasks.

The results in Table~\ref{tab:results} show that our proposed model equipped with the enhanced AdaLN block (block (c)) with 20 steps surpasses all the SOTA models in both datasets except for the FAD score in Landscape where MMD is slightly better. Notably, thanks to the lightweight image encoder, our model has less total parameters compared to others SOTA models. Additionally, we also evaluated our model with 100 and 200 steps, further improving the quality of the generated results. In particular, the improvements w.r.t. the metrics from 100 to 200 steps are negligible, which is consistent with the observations made in \cite{lipman2023flow}.

Regarding A2V and V2A comparisons (Table~\ref{tab:nips-comparison}), we chose to generate the samples using 200 steps. Results in terms of FVD and KVD for A2V  generation are competitive w.r.t. Kim \etal \cite{kim2024a}, yet FAD value is far worse. This is likely because we depend on a vocoder to convert the mel spectogram back to a WAV, while Kim \etal architecture uses a closed-source WAV encoder-decoder architecture \cite{zeghidour2021soundstreamendtoendneuralaudio}. However, since the gap between our generated audio and the corresponding GT is very small (see Table \ref{tab:results}), our model still generates high-quality audio.

\begin{table}[h!]
    \centering
    \adjustbox{width=\linewidth}{
    \begin{tabular}{c|cc?cc|c?cc|c}
        \toprule
        \multicolumn{3}{c?}{} & \multicolumn{3}{c?}{\textbf{AIST++}} & \multicolumn{3}{c}{\textbf{Landscape}} \\ 
        \midrule
        \multicolumn{3}{c?}{} & \multicolumn{2}{c|}{A2V} & V2A & \multicolumn{2}{c|}{A2V} & V2A \\ 
        \midrule
        Model  & Params & Steps & FVD $\downarrow$ & KVD $\downarrow$ & FAD $\downarrow$ & FVD $\downarrow$ & KVD $\downarrow$ & FAD $\downarrow$ \\ 
        \midrule
        Kim \etal $^*$ \cite{kim2024a} & 731 M & 250 & 38.04 & 5.27 & 1.10 & 86.79 & 4.30 & 0.78 \\
        \customrule
        \textbf{\arch{}} & 421 M & 200 & 46.81 & 8.72 & 8.52 & 94.47 & 4.00 & 9.34\\

        \bottomrule
    \end{tabular}
    }
    \caption{Our model compared with Kim \etal \cite{kim2024a} in A2V and V2A generation. $^*$Metrics of the model are taken from the paper since \textbf{neither code nor checkpoints are available.}}
    \label{tab:nips-comparison}
\end{table}

Finally, we evaluated our model through a user study on AIST++, with subjects randomly sampled from experts and non-experts in computer vision. Participants are asked to choose the best video (among a pair) in terms of AV quality and alignment. We compared our model with MMD and asked the participants to evaluate 10 randomly sampled pairs of videos per form. We also compared with Kim \etal model. However, since their model is closed-source and we could not sample new videos, we compared our generated videos with videos downloaded from their project website\footnote{\url{https://avdit2024.github.io/}}. The general results are shown in Table.~\ref{tab:qualitative} and prove that our model is considered the best in the majority of cases.

\begin{table}[h]
    \centering
    \begin{tabular}{c?c|c}
    \toprule
    Models & Quality & AV alignment \\
    \midrule
    Ours vs MMD \cite{ruan2023mm}  & 81\% & 78\% \\ 
    Ours vs Kim \etal \cite{kim2024a} & 80\% & 76\%\\
    \bottomrule
    \end{tabular}
    \caption{User study to evaluate the overall quality and the AV align between our models and the current state-of-the-art models.}
    \label{tab:qualitative}
\end{table}


\paragraph{Results on Long Video Generation.}

\begin{figure*}
    \centering
    \includegraphics[width=\linewidth]{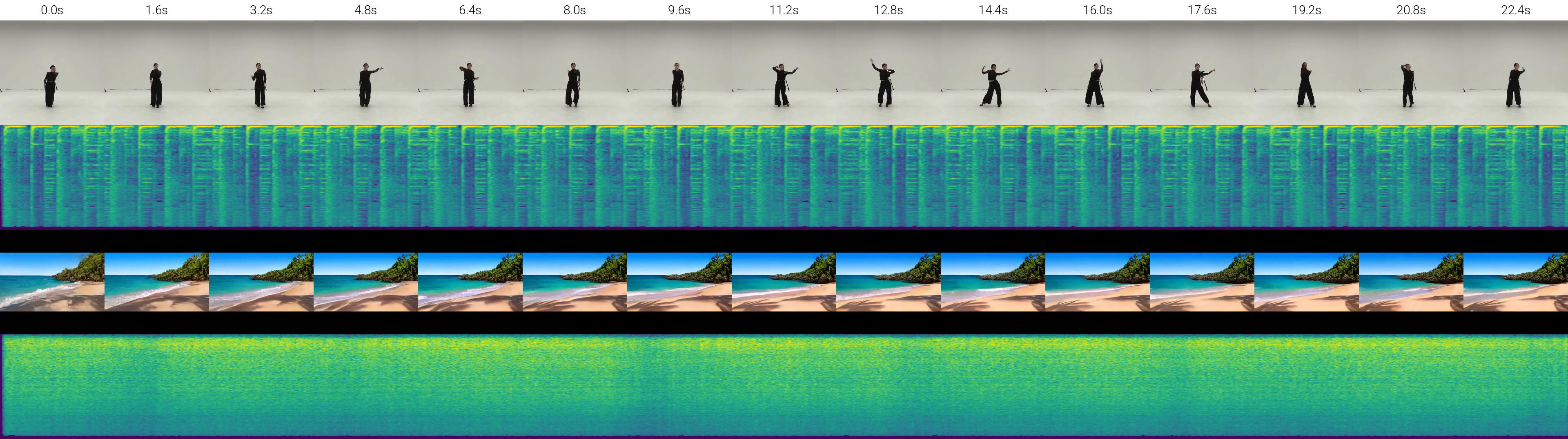}
    \caption{Long videos generated by our model on both AIST++ and landscape. Frames are sampled every 1.6s for visualization.}
    \label{fig:long-video}
\end{figure*}

In Fig.~\ref{fig:long-video} examples of longer videos (240 frames \textit{i.e.}, 24 seconds) are depicted, to show that the quality remains consistent throughout their duration, without visible signs of degradation. 

\begin{figure}
    \centering
    \includegraphics[width=1\linewidth]{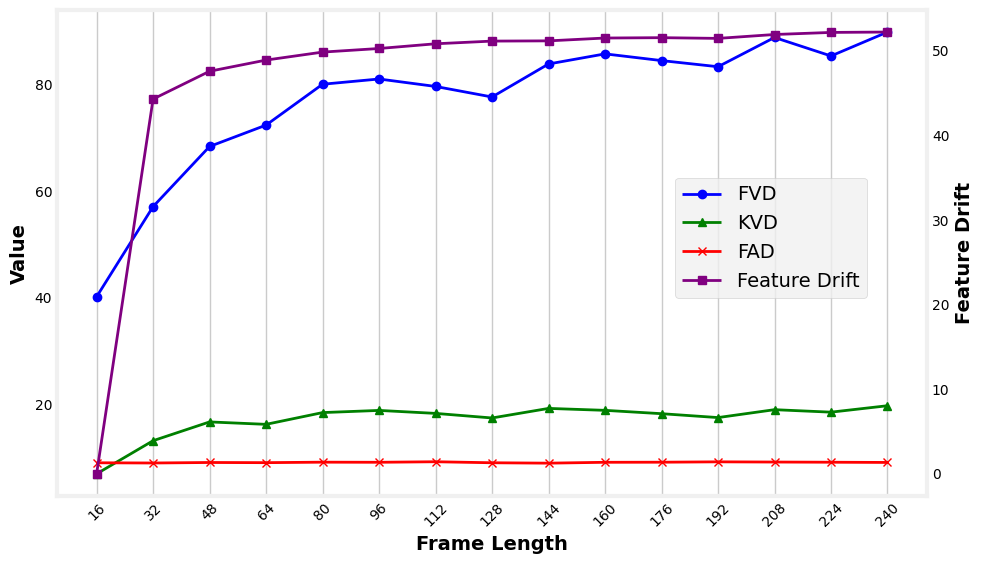}
    \caption{AV metrics and feature drift calculated on long (\textit{i.e.}, 240 frames) generated videos using a sliding window of 16 frames.}
    \label{fig:long_video_metrics}
\end{figure}

To provide a quantitative evaluation of the video quality in the long run, we compute the metrics of Sect.~\ref{subsec:metrics} on the 240 frames-long sequences, using a sliding window of 16 frames. The average results for 2048 videos are shown in Fig. \ref{fig:long_video_metrics}. The figure unexpectedly reveals quite a big jump in the metrics after the first evaluation window. We argue that this is caused by the rolling diffusion design and, particularly, when switching from the pre-rolling to the full rolling phase. Indeed, during the pre-rolling phase, each frame $k < T$ is conditioned only on the previous $k-1$ frames in the sequence, due to temporal attention. This makes the generative process easier since the model needs to take into account less conditioning information. On the contrary, in the full rolling phase, at each diffusion step a fully denoised frame exits the window, while a new one (pure noise) gets in, meaning that each new frame will be conditioned by all the previous frames in the window, equal to $T-1$. Overall, this seems to lead to a slight shift in the generated data distribution, even though no perceivable changes are observed (see the supplementary to check videos).


\begin{figure}[h]
    \centering
    \includegraphics[width=1\linewidth]{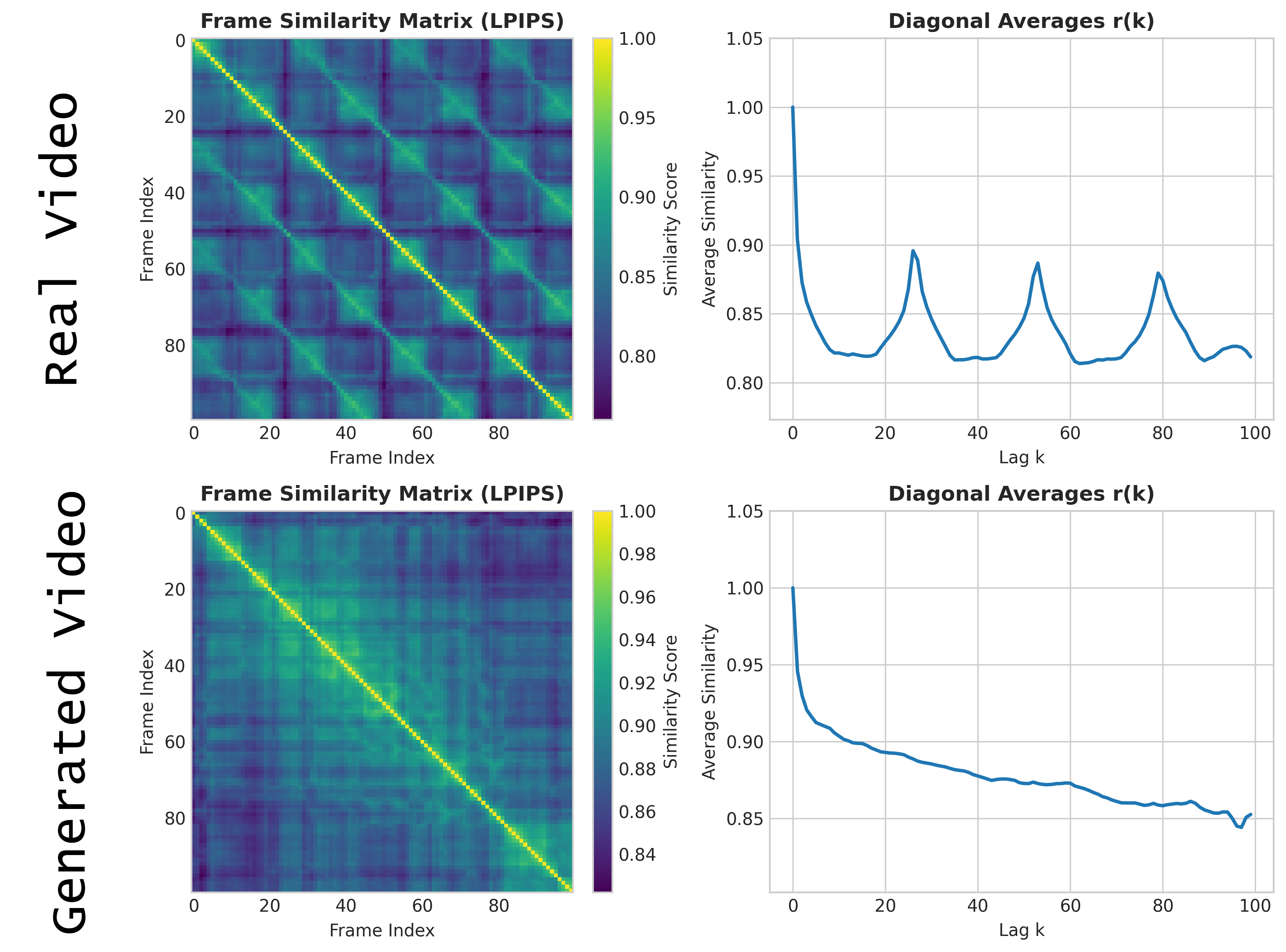}
    \caption{Frame similarity matrix of real vs.\ generated videos from the AIST++ dataset, illustrating the presence of loops in real videos and their absence in generated content.}
    \label{fig:comparison}
\end{figure}

To validate this hypothesis, we calculated the distance between features extracted from the first window (16 frames) of the sequence, and all the other windows. To extract the features we used the same network used to calculate the FVD~\cite{carreira2017quo}. The idea is that, if the above conjecture holds, then we will observe a strong initial drift in the encoded features. From the results (purple line in Fig.~\ref{fig:long_video_metrics}), we indeed observe a large drift after the first chunk. After that though, the drift increment becomes minimal, suggesting that after the transitory phase from pre-rolling to full rolling, the features get stabilized. The same behavior is observed for all the other metrics. Nonetheless, despite the metrics do increase to some extent, they are still comparable or lower than previous state-of-the-art methods, suggesting that the proposed solution is a viable tradeoff between generation quality and consistency throughout time.

This analysis also suggests that the rolling technique can be improved to limit such drift, yet it is fundamental to ensure consistency in the long run.

\paragraph{Detecting Looping Sequences in Videos.}
One may wonder whether the generated long videos are just a looped repetition of shorter sequences which would hinder our claim of infinite AV generation. To clearly show this is not the case and our model effectively learned varying motion dynamics, we designed a specific test. The idea is that if a video contains looped sub-sequences, then similar frames will appear at periodic timesteps. To verify that, we compute a pairwise frame-to-frame similarity using the Learned Perceptual Image Patch Similarity (LPIPS) metric~\cite{zhang2018perceptual} on both real and generated videos from the AIST++ dataset, which features complex motion dynamics. From these, we build a pairwise similarity matrix where the $(i,j)$ entry measures the LPIPS between the frames $i$ and $j$. To quantify looping behavior, we then compute the average of off-diagonal similarity values, denoted as \(r(k)\), where $k$ represents the frame offset or the time lag. Peaks in $r(k)$ values denote a looping sequence of $k$ frames \textit{i.e.} a high similarity peak in $r(k)$ means that a general frame $i$ closely resembles the frame $(i+k)$, indicating a repetitive pattern of period $k$. 

Fig.~\ref{fig:comparison} illustrates the process for two videos, real and generated. On the left are the two similarity matrices and on the right the $r(k)$ calculated from them. It is evident from the similarity matrix that the real video presents a peculiar diagonal pattern, which is a clear hint of a loop. This can be verified by observing the AIST++ videos which depict people performing repetitive dance moves. The real video presents 3 peaks in $r(k)$, denoting a looping sequence with a length of $\sim 23$ frames. On the contrary, the analyzed generated video does not show such behavior. 

For the sake of completeness, we analyze the whole training split of AIST++, which contains 980 videos, with an average duration of 10 seconds. Among them, we identified 692 videos containing looping sequences. To do so, first we compute the Fourier transform of $r(k)$. Then, in frequency domain, we can identify dominant frequency components and selecting the videos for which the largest component (indicating a loop) is above a fixed threshold. In contrast, when looking at 980 synthetic videos of the same length generated by our model, only 264 exhibited looping behavior. This demonstrates that our model can generate extended video sequences without repetitive loops.


\section{Conclusion}
In this paper, we propose \arch, a novel architecture based on flow matching that takes advantage of a rolling diffusion approach to generate coherent and infinite video samples with synchronized audio tracks and unfixed duration. Moreover, we designed a novel temporal fusion module that enables efficient AV generation, while only requiring a limited amount of frames during training. 
Through a series of experiments and ablation studies on two common datasets, we demonstrate, both qualitatively and quantitatively, our improvements over state-of-the-art methods.

Despite the substantial contribution of the proposed architecture w.r.t. the quality and length of the generated AV sequences, when dealing with such a task, a series of technical challenges remain to be faced. In particular, when trained on AIST++, our model occasionally struggles to generate complex limb movements. Furthermore, when generating long videos, it may sometimes fail to preserve time consistency when some visual elements are obscured beyond the model processing window. In AIST++, this happens in scenarios with partial occlusion or unconventional movements, such as intricate dance sequences. For example, a dancer's arm could cover the logo on their t-shirt for several frames and therefore the model will forget about it. On the other side, in Landscape, in sea scenarios, the water could obscure for an extended duration visual elements that will be then forgotten by the model. This reveals areas for future algorithmic refinement.

{
    \small
    \bibliographystyle{ieeenat_fullname}
    \bibliography{main}
}


\end{document}